# TEFormer: Texture-Aware and Edge-Guided Transformer for Semantic Segmentation of Urban Remote Sensing Images

Guoyu Zhou, Jing Zhang, *Member, IEEE*, Yi Yan, Hui Zhang, *Member, IEEE*, and Li Zhuo, *Member, IEEE*

***Abstract*—Accurate semantic segmentation of urban remote sensing images (URSIs) is essential for urban planning and environmental monitoring. However, it remains challenging due to the subtle texture differences and similar spatial structures among geospatial objects, which cause semantic ambiguity and misclassification. Additional complexities arise from irregular object shapes, blurred boundaries, and overlapping spatial distributions of objects, resulting in diverse and intricate edge morphologies. To address these issues, we propose TEFormer, a texture-aware and edge-guided Transformer. Our model features a texture-aware module (TaM) in the encoder to capture fine-grained texture distinctions between visually similar categories, thereby enhancing semantic discrimination. The decoder incorporates an edge-guided tri-branch decoder (Eg3Head) to preserve local edges and details while maintaining multiscale context-awareness. Finally, an edge-guided feature fusion module (EgFFM) effectively integrates contextual, detail, and edge information to achieve refined semantic segmentation. Extensive evaluation demonstrates that TEFormer yields mIoU scores of 88.57% on Potsdam and 81.46% on Vaihingen, exceeding the next best methods by 0.73% and 0.22%. On the LoveDA dataset, it secures the second position with an overall mIoU of 53.55%, trailing the optimal performance by a narrow margin of 0.19%.**

## I. Introduction

SEMANTIC segmentation of urban remote sensing images (URSIs) involves assigning semantic labels to individual pixels based on image content, such as buildings, vehicles, and trees [1]. This task is fundamental to various applications, including urban planning, environmental monitoring, and disaster response. Compared with general RSI, URSI is characterized by densely distributed objects, complex spatial structures, and a wide range of object scales. These characteristics introduce higher accuracy demands and pose significant challenges for semantic modeling.

Current mainstream segmentation methods can be broadly categorized into Convolutional neural networks (CNNs)-based and Transformer-based approaches. CNNs have been widely used in early RSI segmentation for their strong local feature extraction ability. Li et al. [2] proposed a lightweight pure convolutional module that captures both global context and local textures. However, the limited receptive field makes CNNs less effective in modeling long-range dependencies. Transformer-based architectures address this by capturing global context, yet they still struggle with complex category distributions and blurred boundaries common in RSIs. To address these limitations, various improvements have been explored, such as global-local attention mechanisms [3], detail-structure preservation modules [4], spatial-specific Transformers [5], and hybrid CNN-Transformer frameworks [6]. Our previous work D$^2$SFormer [7] combined cross-shaped window self-attention with convolutional channel attention to jointly model global and local features, achieving better URSI segmentation performance. However, it still lacks explicit modeling of texture and boundary information, leaving room for further enhancement.

In URSI, since objects with similar spatial layouts often differ in texture and edge feature, such fine-grained details are critical for accurate segmentation. Texture features describe surface properties such as roughness and regularity. Previous studies, including the texture enhancement attention module [8] and the quantization and counting operator (QCO) [9], have shown that leveraging texture can significantly improve category discrimination. Notably, the dual-path network proposed by Li et al. [2], which jointly models contextual and local texture information, offers valuable insights for our design. On the other hand, edge features provide contour information crucial for boundary localization. Techniques such as dynamic hybrid gradient convolution [10] and three-branch architectures [11] have proven effective in improving edge prediction. In URSIs, boundary pixels often lie at interfaces between semantic classes and are prone to misclassification. Thus, precise extraction of edge features is essential for improving segmentation accuracy.

Some geospatial objects in URSI exhibit similar spatial shapes but possess subtle differences in texture, which can serve as key cues for distinguishing them. Existing Transformer-based models, such as D$^2$SFormer and BiFormer, mainly focus on enhancing multi-scale representation and attention efficiency but pay limited attention to fine-grained texture cues and edge modeling, resulting in insufficient discrimination among visually similar objects and poor boundary delineation. To this end, we propose a texture-aware and edge-guided Transformer (TEFormer). In the encoder, we

This work was supported in part by the Beijing Natural Science Foundation under Grant L247025. *(Corresponding author: Jing Zhang, Hui Zhang).*

All authors are with the School of Information Science and Technology, Beijing University of Technology, Beijing 100124, China, and Jing Zhang, Li Zhuo are also with the Beijing Key Laboratory of Computational Intelligence and Intelligent System, Beijing University of Technology, Beijing 100124, China (e-mail: zhouguoyu@emails.bjut.edu.cn; zhj@bjut.edu.cn; yanyi1581@emails.bjut.edu.cn; huizhang@bjut.edu.cn; zhuoli@bjut.edu.cn).

use the texture-aware module (TaM) to perceive textural differences among similar objects and mine fine-grained texture features. Meanwhile, the frequent occurrence and co-occurrence of irregular shapes, blurred boundaries, and overlapping spatial distributions among semantic objects result in highly complex and diverse edge morphologies at the pixel level. To handle this, we propose an edge-guided tri-branch decoder (Eg3Head) and introduce an edge-guided feature fusion module (EgFFM) that effectively integrates edge information into semantic feature learning, thereby simultaneously enhancing both the overall segmentation accuracy and edge refinement. The main contributions of this work are summarized as follows:

- Unlike D$^2$SFormer and BiFormer, we integrate texture-awareness and edge-guided mechanisms to jointly model local texture and global contextual features, thereby enhancing the discrimination of geospatial objects with similar spatial structures.
- TaM is introduced for fine-grained texture modeling, which improves sensitivity to subtle texture variations and compensates for the texture-insensitivity of existing Transformer encoders.
- An edge-guided three-branch structure, PASPPM, and EgFFM are combined to capture boundary detail and contextual information to guide multi-feature fusion for accurate URSI semantic segmentation.

## II. Methodology

### A. Overall

TEFormer consists of a texture-aware encoder and the Eg3Head, as illustrated in **Fig. 1(a)**. TaM is applied in the first two encoder stages to enhance texture features, while the last two stages use the dual-attention Transformer encoder from D$^2$SFormer [7] to capture global and local context. The decoder adopts edge, detail, and context branches, whose features are fused via EgFFM and combined with shallow features to form the final representation. An MLP and softmax produce the output, and dynamic upsampling helps recover object boundaries.

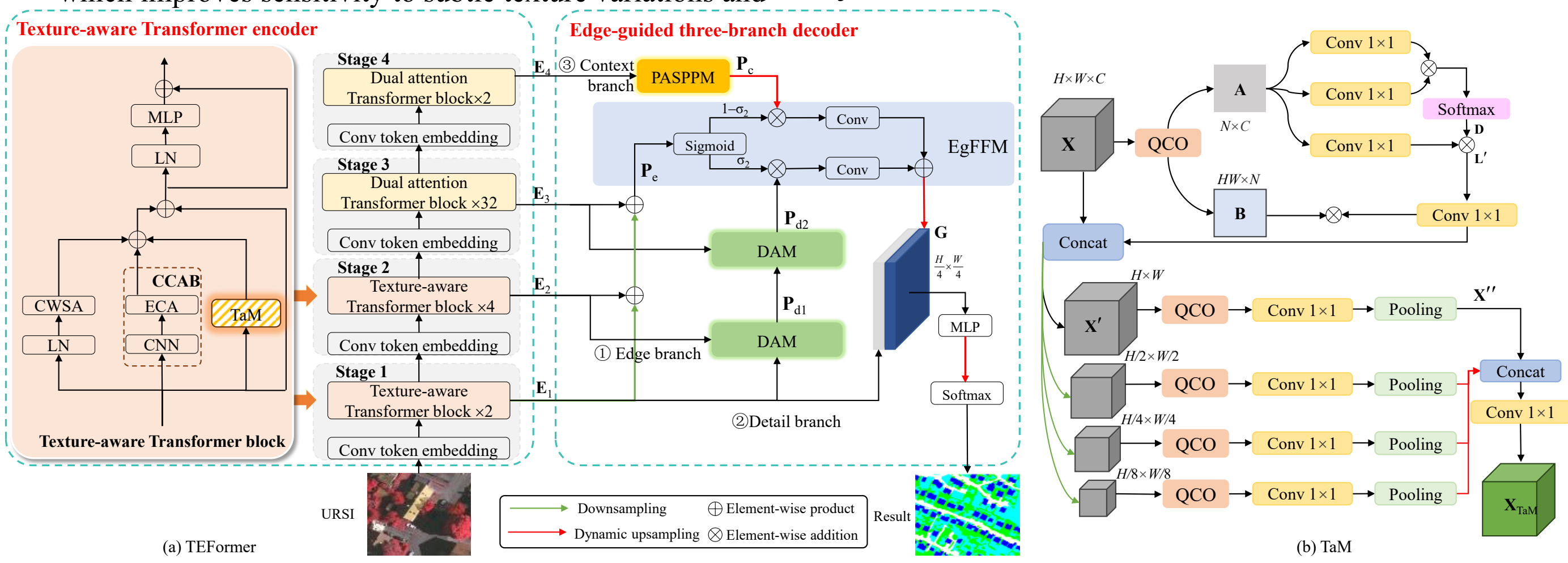

(a) TEFormer (b) TaM

**Fig. 1.** The overall architecture of TEFormer and the detailed structure of TaM.

### B. Texture-aware Module

Conventional convolutions excel at local edges but struggle with the complex textures and distributed spectral features in URSIs. To address this, we introduce a QCO that constructs a texture descriptor via cosine similarity computation, quantization, counting, and encoding, enhancing global texture modeling with low computational cost. Input features are discretized into $L$ levels, and quantized counting features are generated by counting and encoding each level.

However, directly fusing QCO features may result in spatial misalignment or low-order degradation. To address this, we design the TaM, which integrates QCO with an attention mechanism to focus on texture-similar regions. In addition, a multi-scale QCO is employed to reassign quantization levels, further enhancing texture contrast. The structure of TaM is illustrated in **Fig. 1(b)**.

Given the input feature $\mathbf{X}\in\mathbb{R}^{H\times W\times C}$, the module proceeds as follows:

(1) The QCO is applied to quantify the feature distribution, producing the quantized counting feature $\mathbf{A}\in\mathbb{R}^{N\times C}$ and an N-level quantization encoding matrix $\mathbf{B}\in\mathbb{R}^{HW\times N}$.

(2) A graph adjacency matrix $\mathbf{D}\in\mathbb{R}^{N\times N}$ is constructed following the attention mechanism to update the quantization levels $\mathbf{L}'\in\mathbb{R}^{N\times C}$:

$$\mathbf{L}'=\mathrm{Conv}(\mathbf{A})\cdot\mathrm{Softmax}\left(\mathrm{Conv}(\mathbf{A})^{\mathrm{T}}\cdot\mathrm{Conv}(\mathbf{A})\right). \tag{1}$$

(3) After combining $\mathbf{L}'$ with $\mathbf{B}$, they are concatenated with $\mathbf{X}$ to generate the enhanced feature $\mathbf{X}'$:

$$\mathbf{X}'=\mathrm{Concat}\left(\mathrm{Conv}(\mathbf{L}')\cdot\mathbf{B},\mathbf{X}\right). \tag{2}$$

(4) A single QCO cannot capture the multi-scale and continuous texture variations in URSIs, therefore TaM uses multiple QCO branches. The feature map $\mathbf{X}'$ is processed by four branches at distinct scales. Each branch re-quantizes textures at a specific scale, and convolution plus pooling layers align and aggregate the outputs, producing a semantically enriched, multi-scale feature map $\mathbf{X}''$:

$$\mathbf{X}''=\mathrm{Pool}\left(\mathrm{Conv}\left(\mathrm{QCO}(\mathbf{X}')\right)\right). \tag{3}$$

(5) The features from all four scales are dynamically upsampled and concatenated to fuse into the texture features

$\mathbf{X}_{\mathrm{TaM}}$:

$$\mathbf{X}_{\mathrm{TaM}}=\mathrm{Conv}\left(\mathrm{Concat}\left(\mathrm{dys}\left(\mathbf{X}''_{1/8}\right),\mathrm{dys}\left(\mathbf{X}''_{1/4}\right),\mathrm{dys}\left(\mathbf{X}''_{1/2}\right),\mathbf{X}''\right)\right). \quad (4)$$

*C. Edge-Guided Three-Branch Decoder*

To obtain more comprehensive feature representations, we propose a lightweight Eg3Head, which comprises an edge branch, a detail branch, and a context branch.

The edge branch is to retain more complete and detailed object boundaries. Multi-level features extracted by the encoder are first downsampled and summed sequentially to generate the edge feature map $\mathbf{P}_e$. In the detail branch, DAM is introduced to selectively learn semantic representations from features $\mathbf{E}_1$ to $\mathbf{E}_3$, thereby preserving more local structural information. The context branch employs PASPPM to aggregate multi-scale contextual features. This module combines long-range dependencies modeled via average pooling with the large receptive field of dilated convolutions, effectively representing features at spatial scales. Residual connections are used within PASPPM to balance new feature learning with the preservation of original information. The detailed structure of PASPPM is shown in **Fig.2(a)**, where PL and ACL denote the Pooling Layer (PL) and the Atrous Convolutional Layer (ACL), respectively.

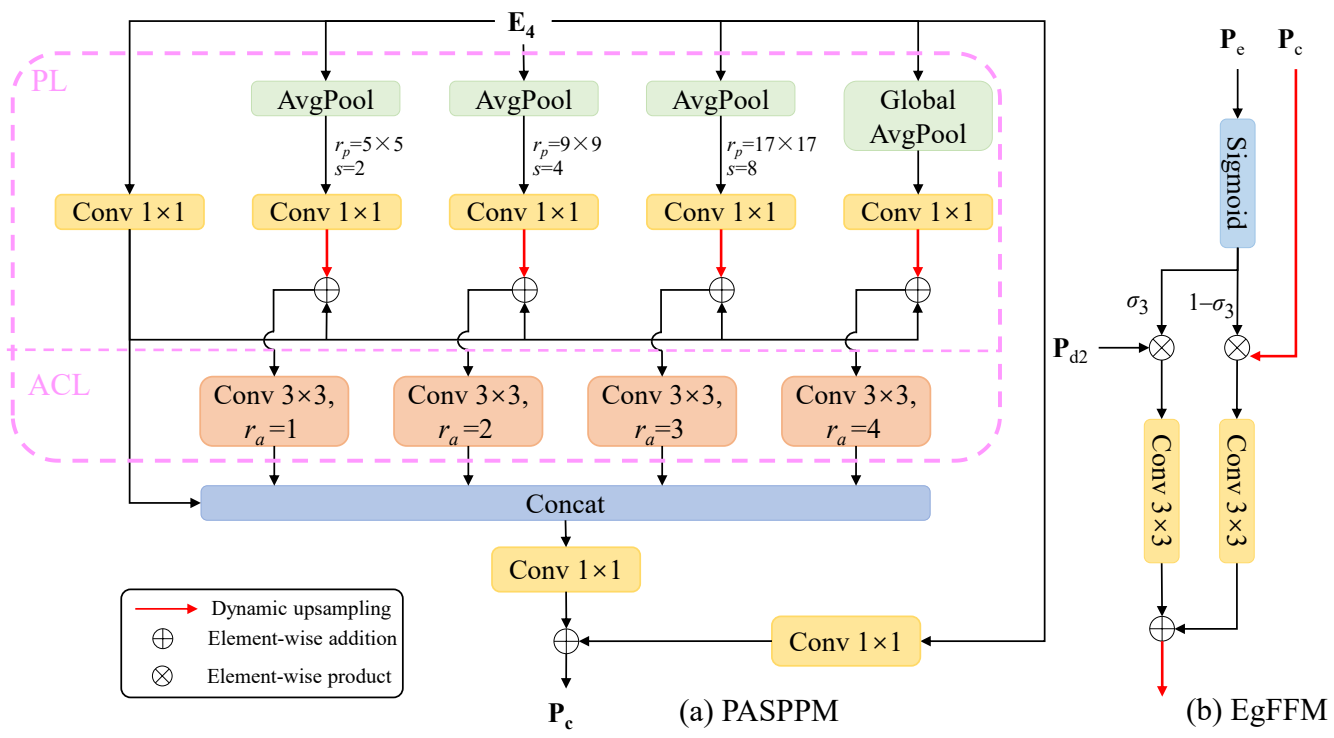

**Fig. 2.** The structure of PASPPM and EgFFM.

*D. Edge-guided Feature Fusion Module*

The context branch provides richer semantic representations but often loses spatial details. In contrast, the detail branch retains edge and texture information but lacks high-level semantic guidance and is prone to artifacts. To effectively combine their strengths, we design the EgFFM, as illustrated in **Fig. 2(b)**. EgFFM leverages boundary information to adaptively regulate the contribution of each branch, enabling more balanced and flexible feature fusion.

Let the feature vectors of the corresponding pixels in the $\mathbf{P}_e$, $\mathbf{P}_c$, and $\mathbf{P}_{d2}$ feature maps be denoted as $v_e$, $v_c$, and $v_{d2}$, respectively. Then, the outputs of the Sigmoid function and the EgFFM are given by:

$$\sigma_3=\mathrm{Sigmoid}\left(\upsilon_e\right). \quad (5)$$

Here, $\sigma_3$ represents the output value of each pixel from the $\mathbf{P}_e$ feature map after the Sigmoid function, which confines the outputs to the range (0, 1). A value of $\sigma_3$=0.5 corresponds to the symmetric operating point of the Sigmoid function, indicating an equal weighting between detail and contextual features. It is important to note that $\sigma_3$ is an attention weight learned directly by the network, not a manually specified parameter, and therefore requires no pre-defined threshold.

$$\mathrm{F}_{\mathrm{EgFFM}}=\mathrm{dys}\left(\mathrm{Conv}\left(\sigma_3\cdot\upsilon_{d2}\right)+\mathrm{Conv}\left(\left(1\text{-}\sigma_3\right)\cdot\mathrm{dys}\left(\upsilon_c\right)\right)\right). \quad (6)$$

When $\sigma_3$>0.5, the pixel is identified as belonging to a boundary region, and the module accordingly assigns greater weight to detail features to accurately delineate edges; otherwise, contextual features are enhanced to fill the interior of objects. This dynamic weighting enables the EgFFM to adaptively balance the contributions from different branches, improving both feature integration and ultimate segmentation performance on URSI.

## III. EXPERIMENTS

*A. Dataset*

We evaluate the performance of our TEFormer on the ISPRS Potsdam/Vaihingen dataset [12], and the LoveDA dataset [13]. These datasets contain diverse urban object categories (e.g., car and building) and are widely utilized for semantic segmentation in URSIs. Specifically, for the LoveDA dataset, we perform a focused evaluation on five major urban categories: background, building, road, water, and barren.

*1)* **Potsdam** dataset contains 38 samples with a resolution of 6000×6000 pixels. It includes six land cover classes: impervious surface, tree, building, low vegetation, car, and background. In our experiments, the dataset was randomly split into training, validation, and test sets at a ratio of 7:1:2.

*2)* **Vaihingen** dataset consists of 33 samples with an average size of 2494×2064 pixels, including near-infrared (NIR), red (R), and green (G) bands. It shares the same categories as Potsdam, and uses the same dataset split proportions.

*3)* **LoveDA** dataset includes 5,987 high-resolution remote sensing images, each sized 1024×1024 pixels. It contains seven land cover classes: background, building, road, water, barren, forest, and agriculture. Following the official split, we used 2,522 images for training, 1,669 for validation, and 1,796 for testing.

*B. Implementation Details*

Experiments were conducted in PyTorch on a TITAN Xp GPU. We adopted a comprehensive data augmentation strategy that includes random scaling, rotation, horizontal and vertical flipping, as well as photometric perturbations. Image patches of size 512×512 were extracted using a sliding window approach. The model was trained for 160k iterations with AdamW (lr=6e-5, weight decay=0.01, batch size=2) on 512×512 image patches. For evaluation, we adopted mean Intersection over Union (mIoU), mean F1 score (mF1), and pixel accuracy (PA). Model complexity was measured in terms of parameter count (Params), floating-point operations per second (FLOPs), and test time.

*C. Comparison with Other Methods*

To evaluate our method, we conducted comparisons with several mainstream semantic segmentation methods (see **Table I**) including the efficient CNN-Transformer hybrid FSegNet [14], the lightweight networks PIDNet [11] and BiFormer [15], as well as CMTFNet [16], Spatial-specificT [5], ESST [17],

TransNeXt [18], and D$^2$SFormer [7]. All of these methods are available for deployment on a TITAN Xp GPU. The visualization results are shown in **Fig. 3**.

*1) Potsdam (Row 1):* TEFormer achieves the best performance, with an mIoU of 88.57%, mF1 of 94.37%, PA of 94.98%. Compared to D$^2$SFormer, TEFormer yields an mIoU improvement of 0.73%. These gains are attributed to the integration of TaM in the encoder and Eg3Head, which together enhance the model's ability to capture subtle texture variations and leverage edge cues to guide the fusion of contextual and detailed features. Compared to other methods, TEFormer also benefits from PASPPM's large receptive field for effective multi-scale context aggregation. In the enlarged region, only TEFormer correctly delineates the boundary between low vegetation and the background, whereas all other methods exhibit blurring boundaries or misclassification.

*2) Vaihingen (Row 2):* TEFormer reaches the best results across multiple key metrics, including 81.46% mIoU, 89.24% mF1, and 90.64% PA. Its texture-aware encoder enhances discrimination of similar geospatial objects, while the Eg3Head effectively fuses contextual, detailed, and edge features for accurate segmentation. Compared to other models, TEFormer achieves notably better results. Notably, TEFormer accurately segments the curved impervious surfaces, surrounding the low vegetation. Other methods produce fragmented outputs or only vaguely identify the low vegetation areas.

*3) LoveDA (Row 3):* While its mIoU is slightly below BiFormer, TEFormer notably outperforms D$^2$SFormer and achieves competitive results, particularly in building and road categories, reflecting stronger land cover recognition. This improvement stems from its edge-guided design, which enhances boundary modeling and reduces local ambiguity. Within this intricate scene, TEFormer is the only method that provides superior segmentation in the boxed region. Other approaches either fail to detect the background or capture only very limited parts.

*4) Complexity analysis*: Although TEFormer exhibits higher complexity than CMTFNet and D$^2$SFormer due to TaM and Eg3Head, with 52.67M Params, 72.25G FLOPs, and a test time of 0.101s, it achieves the highest segmentation accuracy. These results indicate that TEFormer maintains a favorable balance between performance and efficiency.

TABLE I

COMPARISON WITH THE SOTA METHODS ON THREE DATASETS

| Method | Potsdam | | | Vaihingen | | | LoveDA | | | | | | Params (M) | FLOPs (G) | Test time (s) |
|---|---|---|---|---|---|---|---|---|---|---|---|---|---|---|---|
| | mIoU (%) | mF1 (%) | PA (%) | mIoU (%) | mF1 (%) | PA (%) | Back-ground | Build-ing | Road | Water | Barren | mIoU (%) | | | |
| FSegNet [14] | 80.52 | 88.21 | 91.57 | 75.15 | 84.46 | 90.90 | 36.75 | 59.91 | 57.02 | 78.71 | 14.83 | 48.21 | 33.28 | 110.25 | **0.047** |
| BiFormer [15] | 85.28 | 92.89 | 93.34 | 80.93 | 88.89 | 90.36 | 46.3 | 60.0 | **61.02** | **81.02** | 19.44 | **53.74** | 55.22 | 231.27 | 0.109 |
| CMTFNet [16] | 83.57 | 90.93 | 90.77 | 77.95 | 87.42 | 89.24 | 37.27 | 45.95 | 47.61 | 69.92 | 12.14 | 44.13 | 30.07 | **32.85** | 0.075 |
| PIDNet [11] | 86.74 | 93.17 | 93.55 | 80.21 | 88.34 | 89.70 | 46.29 | 58.43 | 60.00 | 78.71 | 17.22 | 52.10 | 37.31 | 34.46 | 0.092 |
| Spatial-specificT [5] | 87.61 | 93.26 | 93.97 | 80.08 | 88.74 | 90.36 | 46.05 | 57.31 | 58.41 | 79.47 | 16.05 | 51.81 | 58.96 | 81.91 | 0.196 |
| ESST [17] | 87.81 | 93.43 | 93.94 | 79.36 | 88.27 | 90.06 | 42.3 | 48.86 | 51.44 | 75.56 | 11.88 | 47.07 | **28.01** | 60.67 | 0.088 |
| TransNeXt [18] | 87.22 | 93.76 | 94.28 | 81.14 | 88.82 | 90.47 | 33.42 | **60.4** | 57.56 | 79.88 | 14.95 | 50.32 | 58.41 | 238.68 | 0.144 |
| D$^2$SFormer [7] | 87.84 | 94.02 | 94.63 | 81.24 | 88.99 | 90.55 | 46.97 | 59.09 | 59.35 | 80.54 | 19.45 | 53.19 | 52.10 | 51.42 | 0.094 |
| TEFormer(ours) | **88.57** | **94.37** | **94.98** | **81.46** | **89.24** | **90.64** | **47.2** | 60.0 | 60.07 | 80.87 | **19.54** | 53.55 | 52.67 | 72.25 | 0.101 |

Note: Bold indicates the best, and underlined indicates the second best.

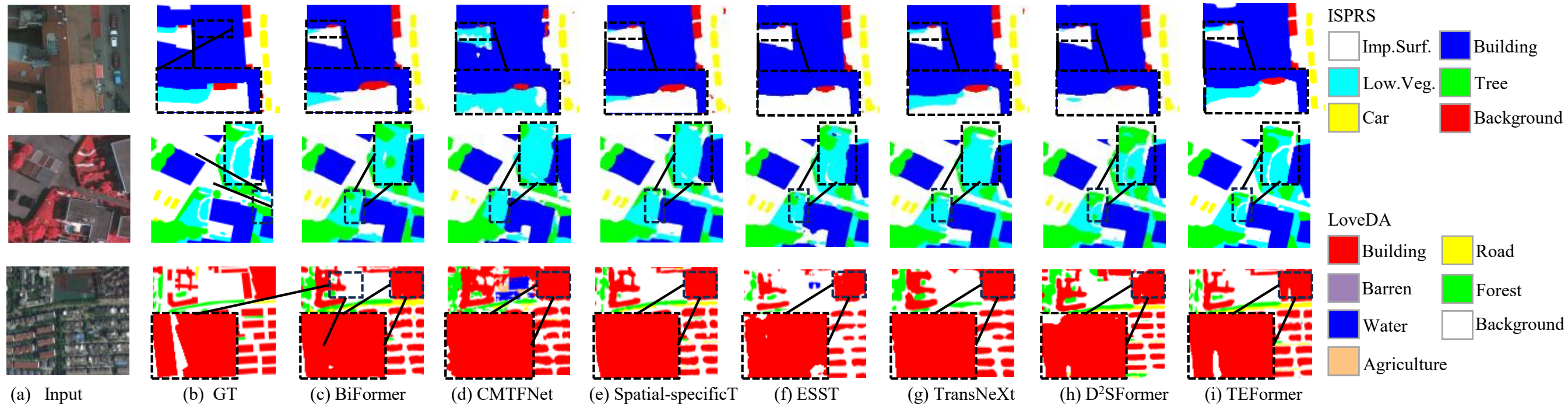

**Fig. 3.** Example of Visualization Results.

### *D. Ablation Studies*

To demonstrate the effectiveness of the proposed method, ablation studies were conducted on the Potsdam dataset focusing on the TaM and the Eg3Head components.

*1) Effectiveness of the TaM:* As shown in **Fig. 4**, using only QCO slightly drops mIoU by 0.05% but increases mF1 by 0.09%, indicating improved texture discriminability but alignment instability due to lack of spatial selection. Attention alone reduces mIoU and mF1 by 0.36% and 0.68%, as it over-focuses on local textures without QCO priors. In contrast, the full TaM combines QCO and attention: QCO quantizes and counts features to establish distinguishable texture levels, offering global texture distribution priors. Attention then spatially selects key regions, guided by these quantized references. This synergy enhances discriminative textures at precise locations, balancing holistic texture recognition with regional discrimination, and boosting mIoU and mF1 to 88.57% and 94.37%, achieving optimal performance.

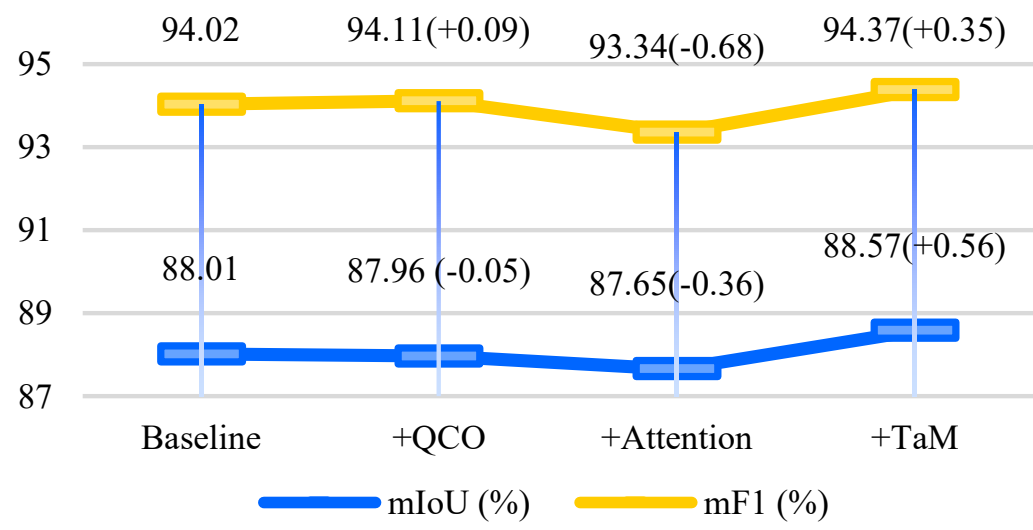

**Fig. 4.** Effectiveness of the TaM.

*2) Effectiveness of the Eg3Head:* To evaluate the effectiveness of Eg3Head, we performed ablation studies on the Potsdam and Vaihingen datasets. As summarized in **Table II**, the first row presents the results of the baseline model ($D^2$SFormer), with the values in parentheses indicate the mIoU improvement relative to the baseline. The experimental results reveal that TEFormer's performance improves consistently as the PASPPM, DAM, and EgFFM are sequentially incorporated. With the full integration of Eg3Head, the model achieves performance gains of 0.73% on Potsdam and 0.22% on Vaihingen, demonstrating robust generalization capability. Moreover, with a controllable increase in parameters and computational cost, the components of Eg3Head effectively enhance feature extraction and fusion.

TABLE II
EFFECTIVENESS OF EG3HEAD

| PASPPM | DAM | EgFFM | mIoU (%) | | M-Params | G-FLOP |
|---|---|---|---|---|---|---|
| | | | Potsdam | Vaihingen | | |
| | | | *87.84* | *81.24* | *52.10* | ***51.42*** |
| | | | 87.21(-0.63) | 81.07(-0.17) | **51.37** | 69.41 |
| ✓ | | | 87.82(-0.02) | 81.32(+0.08) | 52.50 | 69.57 |
| | ✓ | | 87.78(-0.06) | 81.21(-0.03) | 51.40 | 69.67 |
| | ✓ | ✓ | 87.88(+0.04) | 81.44(+0.20) | 51.55 | 72.09 |
| ✓ | ✓ | | 88.20(+0.36) | 81.36(+0.12) | 52.52 | 69.83 |
| ✓ | ✓ | ✓ | **88.57**(+0.73) | **81.46**(+0.22) | 52.67 | 72.25 |

*3) Effectiveness of the TaM at different stages:* As shown in **Table III**, placing the TaM in Stages 1 and 2 yields the largest mIoU improvement (0.46%), while the gains in Stages 3 and 4 are considerably smaller compared to Stages 1 and 2 (0.14% and 0.07%, respectively), with the model reaching a maximum mIoU of 88.78%. This indicates that TaM is more effective for extracting low-level texture features, and thus we embed it only in the first two stages to balance performance and efficiency.

TABLE III
EFFECTIVENESS OF TAM AT DIFFERENT STAGES

| Stage 1 | Stage 2 | Stage 3 | Stage 4 | mIoU (%) |
|---|---|---|---|---|
| ✓ | | | | 88.11 |
| ✓ | ✓ | | | 88.57(+0.46) |
| ✓ | ✓ | ✓ | | 88.71(+0.60) |
| ✓ | ✓ | ✓ | ✓ | **88.78**(+0.67) |

## IV. CONCLUSION

To address the challenges of texture similarity and blurred boundaries in URSI semantic segmentation, we propose TEFormer that integrates texture-aware and edge-guided mechanisms. TEFormer enhances regional discriminability and boundary representation by leveraging texture features, edge-guided decoding, and multi-scale fusion. It outperforms some comparative methods across multiple datasets, and is competitive in detail and edge segmentation. For future work, we will explore lightweight model designs and multi-source data integration to further improve generalization capability.